# Data-Driven Hints in Intelligent Tutoring Systems

Sutapa Dey Tithi[0009-0007-5815-882X]1, Kimia Fazeli[0009-0001-7773-7161]1, Dmitri Droujkov[0009-0008-3561-7509]1, Tahreem Yasir[0000-0001-6505-1915]1, Xiaoyi Tian[0000-0002-5045-0136]1, Tiffany Barnes[0000-0002-6500-9976]1

[1]North Carolina State University, North Carolina, USA

stithi@ncsu.edu, kfazeli@ncsu.edu, ddroujk@ncsu.edu, tyasir@ncsu.edu, xtian9@ncsu.edu, tmbarnes@ncsu.edu

**Abstract:** This chapter explores the evolution of data-driven hint generation for intelligent tutoring systems (ITS). The Hint Factory and Interaction Networks have enabled the generation of next-step hints, waypoints, and strategic subgoals from historical student data. Data-driven techniques have also enabled systems to find the right time to provide hints. We explore further potential data-driven adaptations for problem solving based on behavioral problem solving data and the integration of Large Language Models (LLMs).

**Definition(s):**

Interaction Network: A graph-based representation capturing student solution paths, where nodes represent problem states and edges represent actions taken by students to transition between states.

Markov Decision Process (MDP): A mathematical framework used to model decision-making. In ITS, MDPs are applied to interaction networks to determine the most effective sequence of hints based on historical student success.

Hint: A form of instructional support designed to nudge a learner's reasoning in a productive direction, typically delivered on-demand to activate existing knowledge or prompt necessary inferences. Next-step hints suggest the immediate next problem-solving action. Higher-level and subgoal hints provide future targets, helping students decompose problems into manageable chunks.

# Introduction

Intelligent Tutoring Systems (ITS) promote learning through adaptive, personalized support, especially through hints. Historically, experts authored hints based on rules or anticipated student errors. However, this approach faces significant challenges, especially in open-ended problem solving: the expert authoring burden, the inability to predict unconventional student solution paths, and the assistance dilemma: the need to provide enough information without circumventing learning. This entry explores data-driven hint generation to solve these challenges. By leveraging Interaction Networks built from historical solution traces, ITS can automatically derive contextually relevant guidance for most student problem solving states. The objective of this entry is to detail the evolution of these data-driven techniques, from simple next-step hints to more sophisticated strategic scaffolds like Waypoints and Subgoals. We conclude with a discussion of the potential new data-driven techniques and the integration of Large Language Models (LLMs) for scalable personalized supports.

# Hints

Hints are one of the most fundamental forms of instructional support in human learning. Hints serve an important pedagogical purpose: to provide just enough guidance to keep a learner moving forward without removing the productive struggle that drives deeper understanding (VanLehn, 2006; Kapur, 2008). Hints can take various forms, including cues, feedback, opportunities for reflection, or supplementary information, and can be delivered at varying levels of granularity, from specific next-step hints to abstract strategic guidance (VanLehn, 2006). Critically, hints preserve the learner's agency by nudging reasoning in a productive direction rather than replacing it (Koedinger & Aleven, 2007). The balance between telling too much and saying too little, known as the assistance dilemma, has long been recognized as central to effective tutoring (Koedinger & Aleven, 2007). To understand how hints function and evolve, it is useful to categorize them through a pedagogical taxonomy. Hints generally fall into three broad categories: procedural next-step hints, which guide students on the specific execution of a step; strategic or metacognitive hints, which help students understand overarching problem-solving frameworks and goal setting; and bottom-out hints, which simply provide the exact answer or final step when a student is entirely stuck (Hannafin et al., 2013; Roll et al., 2011). The main challenge remains to deliver the right hint to the right student at the right moment, and at scale. Data-driven hints seek to scale hint generation for domains

Data-driven hints

with large solution spaces (e.g., programming), leverage prior student data on timing to predict when hints are needed, and test out hint types for their effectiveness for diverse student needs.

### Next-Step Hints

The foundational application of data-driven hint generation is the next-step hint, a form of procedural guidance. These hints suggest the exact next action a student should take to progress, typically achievable within a single step (Stamper et al., 2008). Next-step hint generation has been successfully applied across various domains (Barnes & Stamper, 2010; Price et al., 2016; Rivers & Koedinger, 2017). These next-step hints significantly reduce immediate cognitive load and improve tutor completion rates (Stamper et al., 2013; Cody et al., 2022)

Despite the benefits of next-step hints, they present long-term pedagogical limitations. Behaviorally, highly granular hints can inadvertently encourage "hint abuse," where students rapidly request hints without meaningfully engaging with the material (Cody et al., 2018). Pedagogically, next-step hints often fail to teach transferable problem-solving strategies. Students may learn what to do next in a specific context, but not the why, potentially failing to foster self-regulated problem-solving skills (Koedinger & Aleven, 2007). Lastly, another common issue is some students deliberately choose not to use hints (i.e., "help avoidance") even when they need it (Maniktala et al., 2020).

### Higher-Level Hints (Waypoints)

Cody et al. (2022) introduced "Waypoints," which are higher-level data-driven hints that project several steps ahead of the student's current state. As students progress from novices to more advanced levels, they benefit from abstract goal representations that foster a deeper understanding of problem structure (Catrambone, 1998; Sweller & Levine, 1982). While Waypoints were found to slightly increase cognitive load for low-proficiency students, they significantly improved problem-solving efficiency for learners with higher prior knowledge.

### Subgoal Hints

Closely related to Waypoints, subgoal hints help students decompose complex tasks into manageable sub-problems (Margulieux et al., 2016). Using data-driven graph-mining techniques, ITSs can identify common intermediate targets derived from expert and successful student data (Eagle & Barnes, 2014; Shabrina et al, 2023). Shabrina et al. (2023) utilized a data-driven technique called Approach

Mining to break problem-solving solutions into meaningful chunks. Guiding students to complete partially-worked examples broken into these data-driven chunks with explicit instruction on backward strategy significantly enhanced students' subgoaling skills and reduced cognitive load.

## Data-Driven Hint Generation: Data Types

Data-driven hint generation algorithms leverage data from prior student attempts at a given problem to provide feedback to new students attempting the same problem (Barnes & Stamper, 2010). The most commonly used data type is *student solution trace* data: the sequence of states and actions that students produce as they work through a problem. The state can be thought of as a snapshot of the current problem solving attempt/progress, and an action is typically something that changes the state, like applying a domain principle to derive a new value or equation. In propositional logic proofs, Barnes and Stamper (2010) represented each state as the set of propositions derived in a student attempt, with each action corresponding to applying a logic rule to derive a new proposition. An interaction network was formed by taking the union of all individual student solution graphs, retaining frequency information (Eagle et al., 2012). In programming tutors, interaction networks or traces consist of sequences of code snapshots. Price et al. (2016) represented code states as abstract syntax trees (ASTs), while Rivers and Koedinger (2017) used *canonicalized* code representations produced through variable anonymization, expression normalization, and dead code elimination.

The second data type is behavioral *interaction features*—temporal (average step time, total time, etc.), performance (correct/incorrect steps, accuracy, solution length, etc.), and help-seeking patterns (hints requested, hint use ratio, etc.) that characterize ITS engagement. Chi et al. (2011) used such features to create reinforcement learning based pedagogical policies to determine problem types. Maniktala et al. (2020) used these features to develop a data-driven help-need predictor to trigger proactive hints in logic. Marwan et al. (2020) used code compilation results, idle time between actions, and code edit patterns to trigger proactive adaptive immediate feedback for novice programmers. Mostafavi et al. (2017) developed data-driven proficiency profiling for the Deep Thought logic tutor by analyzing per-rule success rates and application frequencies across attempts, enabling targeted problem selection and more informed hint delivery. Eagle and Barnes (2014) proposed several new types of data-driven feedback: temporal– leveraging common time patterns to let students know when they are taking longer than expected, hazard– providing alerts when a student is off-track in comparison to others or experts, and high-level hints that could provide intermediate targets beyond next steps.

## Data-Driven Methods Across Domains

In the Logic domain, the *Hint Factory* pioneered data-driven hint generation by applying *Markov Decision Processes* (MDPs) to interaction networks constructed from student solution trace data (Stamper et al., 2008). Barnes and Stamper (2010) demonstrated that this approach could provide correct next-step hints more than 80% of the time across four semesters of logic proof data. In a controlled experiment, Stamper et al. (2013) showed that students using Hint Factory hints attempted and completed significantly more proofs and were less likely to drop out of the tutor. Eagle & Barnes (2014) and Shabrina, et al. (2023) derived Approach Maps by applying graph mining to interaction networks to identify subgoals for more strategic hints. Fossati et al. (2010) adapted the Hint Factory to guide linked-list manipulation in the iList system. Price et al. (2016) adapted the Hint Factory for the iSnap programming tutor. However, programming domains pose scalability challenges because even simple exercises produce enormous numbers of distinct solutions, resulting in vast potential solution spaces (representing all possible reasonable combinations of problem-solving statements) with very sparse observed attempt states (reflecting actual student work). This limits hint coverage (the availability of hints based on very similar prior student attempts) while also introducing uncertainty about which hints might be most aligned to the current student approach. To address sparse state spaces, Paaßen et al. (2017) proposed the *Continuous Hint Factory*, which interpolates between similar states using edit distance. In the ITAP programming tutor, Rivers and Koedinger (2017) addressed sparse state spaces by *canonicalizing* student solutions into a reduced set of representative forms.

Heffernan and Heffernan (2014) developed ASSISTments into a platform for running randomized controlled experiments on hint design, including testing whether hints should be delivered using text or video (Ostrow & Heffernan, 2014) and evaluating crowdsourced feedback/hints from teachers (Patikorn & Heffernan, 2020). Roll et al. (2011) developed the Help Tutor, a metacognitive tutor agent that used a computational model of help-seeking behavior capturing data on effective and ineffective help-seeking patterns to provide real-time feedback on students' help-seeking errors. The Help Tutor improved students' help-seeking behavior, and these improvements transferred to new domain-level content a month later, even without continued metacognitive support. Maniktala et al. (2021) showed that assertions, data-driven hints proactively provided when needed could effectively address help avoidance, the problem of students not asking for help when they need it.

## Large Language Model-Based Hint Generation

Most recently, large language models (LLMs) have emerged as an alternative that does not use historical student data or model solutions. Instead, they leverage prompt engineering, few-shot exemplars, chain-of-thought scaffolding, tool-augmented reasoning (e.g., code execution or symbolic solvers), and self-verification or multi-agent critique pipelines to generate hints directly from the problem statement and the student's current attempt.

LLMs to date have already shown promise in providing quality support for learning programming. Sarsa et al. (2022) demonstrated that OpenAI Codex can automatically generate largely novel and sensible programming exercises and code explanations, although both occasionally contained inaccuracies that required instructor correction. The findings highlight LLMs' promise for scaling content creation in programming education, while underscoring the continued need for human oversight. Roest et al. (2024) found that LLM-generated next-step hints for introductory programming were personalised and contained helpful explanations, but also that hints sometimes contained misleading information and did not provide sufficient detail when students approached the end of the assignment. Phung et al. (2024) demonstrated that prompting LLMs with symbolic information, such as failing test cases and fixed programs, enhanced the quality of generated hints. Tithi et al. (2025) evaluated LLM-generated hints for propositional logic proofs and found that although they achieved 75% accuracy with strong consistency and clarity, they struggled with justification, failed to guide students toward optimal solution paths, potentially limiting deeper learning. Haim et al. (2024) used GPT-3 to generate on-demand explanations in ASSISTments, showing that they outperformed no support and matched teacher-authored explanations, establishing a strong baseline for LLM-generated assistance. Worden et al. (2025) extended this approach by adding a GPT-4 self-verification step, which significantly eliminated error-prone explanations while maintaining learning outcomes, demonstrating the scalability of a generate-and-verify pipeline.

Across these studies, with careful engineering by ITS designers, current LLMs can produce fluent, personalised hints competitive with human-authored content
Data-driven hints

in surface quality, yet struggle with deeper pedagogical demands such as justification, fostering metacognitive thinking, or redirecting students toward optimal solution paths. As LLMs continue to evolve rapidly, the open question shifts from whether LLMs can generate *accurate hints* to whether they can do so in pedagogically intentional ways, raising the need for evaluation frameworks grounded in learning science principles rather than surface quality metrics alone.

## Discussion

The data-driven methods we described above differ in important ways with respect to data requirements, scalability, interpretability, and pedagogical grounding. Solution-space-based methods such as the Hint Factory are interpretable and directly grounded in actual student problem-solving trajectories, making the generated hints concrete and contextually appropriate. However, they require sufficient prior student data to populate the state space and face coverage limitations as domain complexity grows, particularly in programming, where the number of unique solution paths can be very large (Price et al., 2016; Paaßen et al., 2017). Price et al. (2019) systematically compared data-driven hint generation algorithms and found that hint quality can plateau or decrease beyond 15–20 training solutions, and that while student data can outperform a single expert solution, a comprehensive set of expert solutions generally performs best. Extensions such as the Continuous Hint Factory (Paaßen et al., 2017) and code canonicalization (Rivers & Koedinger, 2017) partially address coverage, but add algorithmic complexity and may lose information about individual solution strategies in the process.

Reinforcement learning generated pedagogical policies can drive decisions such as when to give a hint and what type of support to provide (Ausin et al., 2019; Zhou et al., 2019). Their strength lies in adapting to individual student characteristics over time, but they require substantial training data to learn effective policies, can be difficult to interpret, and the learned policies may not generalize well across domains or student populations. RL methods use hint content, whether expert-authored or data-driven, to optimize over action sequences and thus complement the hint generation approaches.

LLM-based approaches can generate hints for any problem without prior student data, model solutions, or domain models (Roest et al., 2024). This makes them particularly attractive for new courses or problems where no historical data exists. However, LLM-generated hints lack the grounding in actual student solution paths that data-driven methods provide, and can produce misleading or incorrect content (Roest et al., 2024). Furthermore, because LLMs do not inherently model student knowledge or learning trajectories, adapting hint content to individual students' current understanding requires additional scaffolding mechanisms. As Denny et al. (2023) have observed, the arrival of LLMs presents both opportunities and challenges for computing education, and understanding how to combine the pedagogical grounding of data-driven methods with the scalability of LLMs remains an open research question.

## Conclusions

Over the last three decades, significant progress has been made in investigating how ITS can scaffold student learning. Beginning with the Hint Factory's application of MDPs to interaction networks, the field has progressed from next-step hints to higher-level strategic supports such as Waypoints and subgoal hints that can foster more transferable problem-solving skills. Recent progress in large language models (LLMs) offers the potential to scale real-time content and feedback generation while enabling more natural, dialogue-based tutor-student interaction (Worden et al., 2025). Despite offering such richer interaction, grounding LLMs on sound pedagogical intent and correctness is still challenging (Tithi et al., 2025). Integrating the rich interaction data and automated discovery of students' problem-solving strategies as well as students' affective and metacognitive states with new LLM techniques holds significant promise for ITS support generation. The techniques we discussed to learn from historical solutions can be further extended to leverage LLMs as dynamic scaffolding agents capable of synthesizing pedagogical patterns across large, heterogeneous student populations. The integration of data-driven techniques with LLMs has significant promise for generation of intelligent supports for non-adaptive learning environments and problem solving domains. Future research should seek how to effectively combine knowledge from the learning sciences to form pedagogically appropriate supports that foster learning and metacognition, techniques from data-driven tutoring to learn from common student approaches, and the natural interaction afforded by LLMs to generate personalized learning supports across problem solving domains.

## Related entries

Design, implementation, and impact of feedback in Artificial Intelligence in Education; Intelligent Tutoring Systems; Learner Modeling; Generative AI in Education; AI in programming education; Adaptive hypermedia

Data-driven hints